\documentclass{article}

\PassOptionsToPackage{numbers,sort}{natbib}
\usepackage[preprint]{neurips_2025}
\bibpunct{[}{]}{,}{n}{}{,}
\setcitestyle{numbers,square,citesep={,}}
\makeatletter
\def\NAT@def@citea{\def\@citea{\NAT@separator}}
\def\NAT@def@citea@space{\def\@citea{\NAT@separator}}
\def\NAT@def@citea@close{\def\@citea{\NAT@@close\NAT@separator}}
\def\NAT@def@citea@box{\def\@citea{\NAT@mbox{\NAT@@close}\NAT@separator}}
\makeatother

\usepackage[utf8]{inputenc}
\usepackage[T1]{fontenc}
\usepackage{hyperref}
\usepackage{url}
\usepackage{booktabs}
\usepackage{amsfonts}
\usepackage{amsmath}
\usepackage{amssymb}
\usepackage{nicefrac}
\usepackage{microtype}
\usepackage{xcolor}
\usepackage{graphicx}
\usepackage{makecell}
\usepackage{subfig}
\usepackage{array}

\title{Digital-to-Physical Transfer of Adversarial Patches for Aerial Vehicle Detection}

\author{%
  Jung Heum Woo \\
  School of Information Technology \\
  Incheon National University \\
  Incheon 22012, Republic of Korea \\
  \texttt{realbb@inu.ac.kr} \\
  \And
  Eun-Kyu Lee \\
  School of Information Technology \\
  Incheon National University \\
  Incheon 22012, Republic of Korea \\
  \texttt{eklee@inu.ac.kr} \\
}

\begin{document}

\maketitle

\begin{abstract}
Deep neural network (DNN)-based object detectors are widely used for analyzing aerial and satellite imagery in applications such as environmental monitoring and urban analytics. Despite their strong performance, these models are known to be vulnerable to adversarial examples, and physical adversarial attacks using printable patterns pose realistic security threats. In this paper, we evaluate physical adversarial patch attacks against an aerial vehicle detector by bridging digital optimization and real-world deployment. Adversarial patches are optimized in the digital domain using a loss function that minimizes the maximum objectness score while incorporating non-printability score (NPS) and total variation (TV) constraints to ensure both printability and spatial smoothness. 
The optimized patches are printed and deployed in three configurations: ON, OFF, and OFF-Side. Experiments using a YOLOv3 detector show that while the OFF patch achieves the highest effectiveness in the digital domain (85.51\% Average Objectness Reduction Rate (AORR)), the ON patch demonstrates superior robustness in physical environments (0.197--0.343 Objectness Score Ratio (OSR)) due to its consistent visibility. Furthermore, our results indicate that weather-based augmentation does not necessarily improve patch optimization in this domain. These findings provide critical insights into the practical vulnerabilities of aerial object detection systems.
\end{abstract}

\section{Introduction}
The rapid growth of aerial and satellite imagery has significantly expanded the use of automated visual analysis systems in a wide range of applications, including environmental monitoring, urban planning, traffic analysis, and disaster response. To efficiently analyze such large-scale image data, deep neural network (DNN)-based object detection models have been widely adopted. Modern object detectors have demonstrated strong performance in identifying objects such as vehicles, buildings, and infrastructure in aerial imagery. Among these models, YOLOv3 has been widely used due to its high detection accuracy and real-time inference capability \cite{ref-yolov3}.

Despite their strong performance, DNN-based detectors are known to be vulnerable to adversarial examples, which are carefully crafted perturbations designed to manipulate model predictions. Previous studies have shown that adversarial perturbations can significantly degrade the performance of both image classification and object detection models. While early research mainly focused on digital adversarial attacks that directly modify pixel values in images, recent work has demonstrated the feasibility of physical adversarial attacks, in which adversarial patterns are printed and placed in real-world environments to influence camera-captured inputs.

However, most existing studies on physical adversarial attacks have focused on ground-based scenarios, such as autonomous driving systems or surveillance cameras. In contrast, aerial and satellite imagery present unique challenges due to large observation distances, varying viewing angles, scale variations, and environmental conditions such as illumination changes and weather effects. As a result, the effectiveness and transferability of physical adversarial attacks in aerial imagery remain relatively underexplored.

Unlike ground-level photography, aerial and satellite imagery is captured through long atmospheric paths, making it susceptible to visibility degradation caused by weather conditions such as fog, haze, and fluctuating illumination. Such environmental variability can significantly undermine the effectiveness of adversarial patches when they are transferred from digital optimization to physical deployment. To address this, we incorporate weather-based augmentation into the patch optimization pipeline to simulate atmospheric effects and environmental noise, thereby investigating its impact on ensuring the robustness of adversarial attacks under diverse real-world capture conditions.

Building on this motivation, this study evaluates physical adversarial patch attacks against an aerial vehicle detector by developing and analyzing a practical attack pipeline that bridges digital optimization and real-world deployment. Adversarial patches are first optimized in the digital domain and then physically printed and deployed in real environments to evaluate their impact on object detection performance. We consider three patch configurations: ON patch placed directly on the vehicle roof, OFF patch placed around the vehicle, and OFF-Side patch in which two patches are placed on both sides of the vehicle. The target detector is pre-trained on the Microsoft COCO dataset \cite{ref-coco} and fine-tuned on the Cars Overhead With Context dataset \cite{ref-cowc} for aerial vehicle detection.

Our experimental results yield several critical insights into the physical robustness of aerial vehicle detectors. First, we found a significant discrepancy between domains: while the OFF patch configuration---which manipulates contextual information---is most effective in digital simulations (achieving up to 85.51\% AORR), the ON patch demonstrates superior effectiveness and robustness in real-world environments. Second, our analysis indicates that weather-based augmentation during the optimization stage does not necessarily improve attack performance and can even hinder the generation of effective adversarial patterns in this specific domain. Finally, we demonstrate that moderate total variation (TV) regularization is essential for balancing the trade-off between physical printability and adversarial potency.
These findings provide insights into the practical vulnerabilities of aerial object detection systems under real-world physical perturbations.

The scientific novelty and main contributions of this paper are summarized as follows:
\begin{itemize}
\item We present a digital-to-physical evaluation of adversarial patch attacks for aerial vehicle detection, focusing on the practical transfer gap between digital optimization and real-world aerial capture.
\item We compare object-level and context-level patch deployment strategies, including ON, OFF, and OFF-Side configurations, and show that the most effective strategy differs between digital and physical domains.
\item We analyze weather-based augmentation and TV regularization in aerial patch optimization, providing empirical evidence that environmental augmentation does not necessarily improve attack effectiveness in this setting.
\end{itemize}

\section{Background and Related Work}
\subsection{Aerial Image Object Detection}
Object detection in aerial and satellite imagery has become an important research topic due to its wide range of applications, including urban analysis, environmental monitoring, and traffic understanding. Compared with natural-image object detection, aerial image detection is more challenging because target objects often appear very small, densely distributed, and visually ambiguous under varying viewpoints and imaging conditions. To address these challenges, convolutional neural network (CNN)-based detectors have been widely adopted in remote sensing tasks.

Among early large-scale applications, Albert et al. \cite{ref-albert} demonstrated that convolutional networks combined with satellite imagery can effectively capture large-scale urban patterns, showing the practical value of deep learning for aerial scene understanding. In addition, Ji et al. \cite{ref-ji} proposed a vehicle detection framework for remote sensing imagery that integrates super-resolution with object detection to improve the detection performance of small vehicles in aerial images. These studies demonstrate that deep learning-based approaches have become the dominant paradigm for aerial object detection.

Recent studies have further improved aerial and remote sensing object detection by addressing the unique characteristics of overhead imagery. Cai et al. \cite{ref-pkinet} proposed PKINet to capture multi-scale local features and long-range contextual information for remote sensing detection. Huang et al. \cite{ref-mutdet} investigated detection pre-training for remote sensing scenes and showed that object-level representation alignment can improve transfer performance, especially under limited annotation settings. These advances indicate that aerial object detectors are becoming increasingly specialized for scale variation, dense object layouts, and arbitrary orientations.

More broadly, the success of deep neural networks in aerial imagery has motivated the development of detection systems for vehicles and other small objects observed from overhead viewpoints. In this context, aerial object detection serves as a critical foundation for many downstream tasks, and its reliability is particularly important when the detection results are used in real-world monitoring or decision-making systems.

Despite their strong performance, aerial image detectors remain vulnerable to distribution shifts caused by viewpoint variation, scale change, illumination, and environmental conditions. These limitations become even more critical in adversarial settings, where carefully designed perturbations can exploit the detector's sensitivity. Therefore, understanding the vulnerability of aerial object detectors is necessary not only from the perspective of model accuracy, but also from the viewpoint of security and robustness.

\subsection{Adversarial Attacks on Object Detection}
Adversarial examples are intentionally crafted perturbations that cause machine learning models to produce incorrect predictions. Goodfellow et al. \cite{ref-goodfellow} established one of the foundational formulations of adversarial examples and showed that deep neural networks are highly vulnerable even to small, human-imperceptible perturbations. Subsequent studies such as DeepFool and Universal Adversarial Perturbations further expanded this line of work by proposing effective optimization-based attack methods and demonstrating that perturbations can generalize across many inputs \cite{ref-deepfool, ref-uap}. Carlini and Wagner \cite{ref-cw} later introduced stronger optimization-based attacks that became a standard benchmark for evaluating model robustness.

While many early studies focused on image classification, later work extended adversarial attacks to object detection, which is a more complex task involving both localization and classification. Eykholt et al. \cite{ref-eykholt} examined attacks on object detectors using adversarial stickers, while Song et al. \cite{ref-song} further discussed physical adversarial examples against object detection models. Chen et al. \cite{ref-shapeshifter} proposed ShapeShifter, a robust physical adversarial attack against Faster R-CNN, showing that detector-specific attacks can remain effective under transformations. Lee and Kolter \cite{ref-lee} also investigated physical adversarial patches for object detection, emphasizing the challenges of patch placement and transferability in detection scenarios.

In the aerial domain, den Hollander et al. \cite{ref-hollander} explored adversarial patch camouflage against aerial detection, suggesting that object detectors operating on overhead imagery are also susceptible to localized perturbations. Similarly, Czaja et al. \cite{ref-czaja} discussed adversarial examples in remote sensing and highlighted the importance of studying adversarial robustness in aerial and satellite imagery. These studies indicate that adversarial attacks on object detection are not limited to ground-based scenes and that overhead detection systems constitute an important yet relatively underexplored attack surface.

\subsection{Physical Adversarial Patch Attacks}
Physical adversarial patch attacks aim to preserve adversarial effectiveness after digital perturbations are transferred into the real world through printing, placement, and image recapture. Kurakin et al. \cite{ref-kurakin} presented one of the earliest demonstrations that adversarial examples can survive the physical imaging process, although their effectiveness may degrade due to viewpoint and environmental variations. To address these limitations, Brown et al. \cite{ref-brown} introduced the adversarial patch paradigm, showing that a localized patch can be optimized to produce universal and robust attack behavior under different transformations. This work became a key foundation for later physical patch attacks.

Subsequent studies improved the realism and robustness of physical adversarial patches by explicitly considering printability, smoothness, and physical-world constraints. Komkov and Petiushko \cite{ref-advhat} proposed AdvHat, in which physically wearable adversarial patterns were designed to attack face recognition systems. In the object detection domain, many studies extended the patch-based attack setting to detection models, demonstrating that adversarial patterns can suppress or alter detections when placed in physically meaningful regions of the scene \cite{ref-eykholt, ref-song, ref-shapeshifter, ref-lee, ref-thys}.

Recent studies have further expanded physical adversarial attacks by considering more realistic deployment factors. Guesmi et al. \cite{ref-dap} proposed a dynamic adversarial patch that maintains a naturalistic appearance while improving robustness to transformations such as clothing creases. Wei et al. \cite{ref-cap} revisited physical patch attacks from the perspective of camera image signal processing and showed that cross-camera variability can strongly affect attack stability. Long et al. \cite{ref-papmot} extended printable patch attacks from frame-level detection to multiple object tracking by disrupting both detection and temporal identity association. Liang et al. \cite{ref-grac} proposed adversarial camouflage that addresses viewpoint conflicts, illumination changes, and texture smoothness for physical object detection evasion. These studies indicate that physical adversarial attacks are increasingly moving toward realistic imaging, motion, and environmental constraints.

Compared with ground-level physical attacks, physical patch attacks in aerial imagery are more challenging because the attack must remain effective across larger observation distances, severe scale changes, viewpoint differences, and environmental factors such as illumination and weather. Literatures provided early evidence that remote sensing models can be attacked using adversarial perturbations \cite{ref-czaja, ref-hollander}. More recent remote-sensing studies have begun to focus specifically on the digital-to-physical transfer problem. Chen et al. \cite{ref-vcoap} addressed visual inconsistency between digitally optimized patches and their physical appearance by incorporating image harmonization into patch generation, while Li et al. \cite{ref-pareto-bg} proposed a physical adversarial background patch for aerial object detection that balances attack effectiveness and patch area using Pareto-efficient optimization. Therefore, physical adversarial patch attacks on aerial object detectors represent an important research direction for understanding the real-world security limitations of overhead vision systems.

\section{Threat Model}
This study considers a threat model in which an attacker attempts to degrade the performance of an aerial vehicle detector using physically deployable adversarial patches. The attacker places adversarial patterns either directly on the target vehicle or around the vehicle so that the detector fails to recognize the object from aerial imagery. The objective of this threat model is to evaluate whether adversarial patches optimized in the digital domain can be effectively transferred to the physical domain and remain effective under real-world aerial capture conditions.

\textit{Adversarial goal:}
The primary objective of the attacker is to reduce the detection confidence of an aerial vehicle detector. Specifically, the attacker aims to minimize the objectness score predicted by the detector so that the target vehicle is either not detected or detected with significantly reduced confidence. By decreasing the objectness score, the attack seeks to suppress the detector's confidence in the presence of the vehicle.

\textit{Knowledge:}
We assume a white-box threat model in which the attacker has full knowledge of the target detection model, including its architecture, parameters, and loss function. Under this assumption, the attacker can optimize adversarial patches directly against the target detector during the digital optimization stage. This setting allows the attack to evaluate the upper bound of adversarial vulnerability under physically realizable patch constraints.

\textit{Capability:}
The attacker is capable of generating adversarial patches in the digital domain and deploying them in the physical world. First, the patch is optimized digitally using the target detector. The optimized patch is then printed as a physical pattern and placed in the scene. In this study, three placement strategies are considered: an ON patch placed on the roof of the target vehicle, an OFF patch positioned around the vehicle, and an OFF-Side patch in which two patches are placed on both sides of the vehicle.

\textit{Strategy:}
During the optimization, the attacker iteratively updates the adversarial patch so as to minimize the objectness score predicted by the detector. By suppressing the objectness score of the target object, the patch reduces the probability that the detector correctly recognizes the vehicle in aerial imagery. This strategy focuses on degrading detection confidence rather than causing misclassification to another object category.

Under this threat model, we evaluate whether digitally optimized adversarial patches can effectively transfer to the physical domain and degrade the performance of aerial object detectors in real-world capture scenarios.

\section{Attack Method}

\subsection{Patch Configuration}

\begin{figure}[htbp]
\centering
\subfloat[\hspace*{1em} \centering ON patch]{%
    \includegraphics[width=0.27\textwidth]{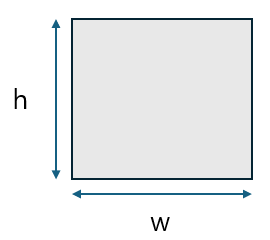}}
\hspace{0.03\textwidth}
\subfloat[\hspace*{1em} \centering OFF patch]{%
    \includegraphics[width=0.27\textwidth]{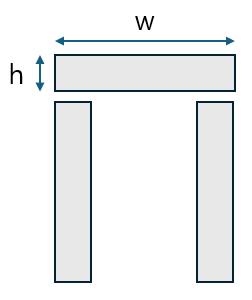}}
\hspace{0.03\textwidth}
\subfloat[\hspace*{1em} \centering OFF-Side patch]{%
    \includegraphics[width=0.27\textwidth]{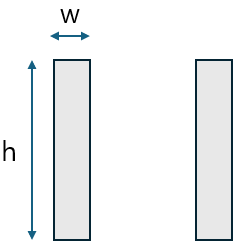}}
\caption{Illustration of the three patch configurations: (\textbf{a}) ON patch, (\textbf{b}) OFF patch, and (\textbf{c}) OFF-Side patch.}
\label{fig:patch_config}
\end{figure}

To evaluate the effectiveness of adversarial patches under different placement strategies, three patch configurations are considered in this study, as illustrated in Figure~\ref{fig:patch_config}. These configurations represent different physical deployment scenarios that may influence the performance of aerial object detectors.

\textit{ON patch:}
The ON patch is placed directly on the roof of the target vehicle. This configuration aims to interfere with the detector by modifying the visual appearance of the object itself. In the digital domain, the ON patch has a size of $200 \times 160$ pixels. For physical deployment, the optimized patch is printed at approximately $1189 \times 841$ mm so that it covers a substantial portion of the vehicle roof and remains visible from aerial viewpoints.

\textit{OFF patch:}
In contrast, the OFF patch is placed around the vehicle rather than directly on it. This configuration attempts to manipulate the contextual information surrounding the target object, thereby affecting the detector's perception of the scene. In the digital domain, the OFF patch is designed as a rectangular strip of size $400 \times 25$ pixels. For physical deployment, the patch is printed at approximately $3200 \times 200$ mm and arranged around the vehicle to influence the detection process.

\textit{OFF-Side patch:}
The OFF-Side patch differs from the OFF patch in both the number and spatial arrangement of the deployed patches. While the OFF patch surrounds the vehicle with three rectangular patches arranged in a $\Pi$-shaped structure, the OFF-Side patch places only two patches on the left and right sides of the vehicle. Similar to the OFF configuration, the OFF-Side patch aims to manipulate the contextual region surrounding the vehicle rather than the vehicle surface itself. By perturbing the side regions of the vehicle, this configuration attempts to influence the detector's contextual perception while using fewer patches than the OFF configuration.

These three patch configurations allow us to investigate how the placement of adversarial patterns affects the robustness of aerial object detectors in both digital simulations and real-world environments. By comparing the ON, OFF, and OFF-Side configurations, we analyze the relative effects of object-level perturbations and contextual perturbations on detection performance.

\subsection{Adversarial Patch Optimization}
\begin{figure}[!t]
\centering
\includegraphics[width=\textwidth]{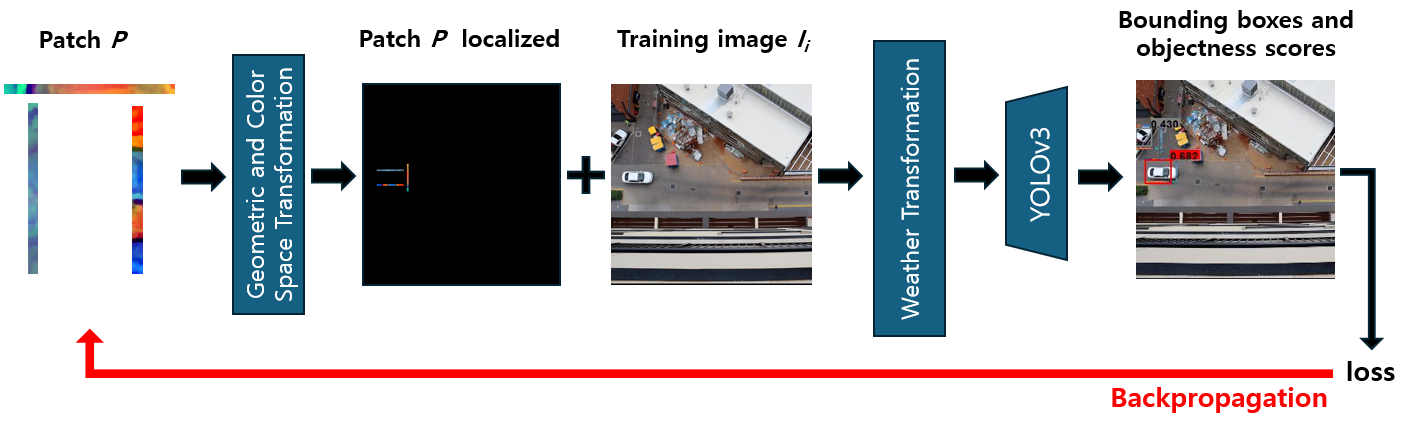}
\caption{Overview of the adversarial patch optimization process.}
\label{fig:patch_optimization}
\end{figure}

The overall pipeline of the adversarial patch optimization process is illustrated in Figure~\ref{fig:patch_optimization}. To generate adversarial patterns that remain effective in real-world environments, the adversarial patches are first optimized in the digital domain before being deployed in the physical domain. The optimization process aims to reduce the confidence of the object detector by minimizing the objectness score of the target object while ensuring that the generated patch can be reliably printed and physically deployed.

Let $P$ denote the adversarial patch applied to an input image $x$. During optimization, the patch is iteratively updated to minimize the maximum objectness score produced by the detector. The optimization objective is defined as follows:
\begin{equation}
L_i(P) = \max(\hat{S}_i^{u}) + \delta \cdot \mathrm{NPS}(P) + \gamma \cdot \mathrm{TV}(P),
\label{eq:patch_loss}
\end{equation}
where $\max(\hat{S}_i^{u})$ encourages the patch to reduce the highest objectness score in the $i$-th image, thereby suppressing the detector's confidence for the target object. The second term, $\mathrm{NPS}(P)$, denotes the Non-Printability Score, which penalizes colors that cannot be reliably reproduced by physical printers. This constraint ensures that the optimized patch remains realizable when printed. The third term, $\mathrm{TV}(P)$, represents total variation regularization, which encourages spatial smoothness by reducing abrupt color changes between neighboring pixels in the patch.

The weighting coefficients $\delta$ and $\gamma$ control the influence of the printability and smoothness constraints, respectively. In our experiments, these parameters are set to $\delta = 0.01$ and $\gamma = 2.5$, following prior work on physical adversarial patches.

In this study, the NPS term is treated primarily as a printability constraint rather than as a main factor for improving adversarial effectiveness. Accordingly, the NPS coefficient is fixed to encourage the optimized patch colors to remain within a physically printable range. In contrast, the TV coefficient is analyzed in more detail because TV regularization directly controls the spatial smoothness of the patch and visibly affects both the optimized patch structure and attack performance. We acknowledge that this setting does not fully reveal the interaction between the NPS and TV terms during optimization; therefore, a joint ablation of both coefficients remains an important direction for future work. The default value $\gamma = 2.5$ is selected because, in our preliminary experiments, $\gamma = 1.0$ and $\gamma = 2.5$ produced patches that appeared less visually unnatural while maintaining competitive attack performance. Among them, $\gamma = 2.5$ provided a favorable trade-off between visual naturalness and adversarial effectiveness.

To improve robustness against environmental variations, several transformations are applied during patch optimization. These transformations simulate changes that may occur during real-world image capture. Specifically, geometric transformations such as random scaling and rotation are applied to account for viewpoint variations. In addition, color-space transformations, including brightness and contrast adjustments, are introduced to simulate illumination changes. Weather-related perturbations are also considered to approximate atmospheric effects such as lighting variations and environmental noise.

Through this optimization process, the generated adversarial patch is designed to remain effective under both digital evaluation and real-world physical deployment conditions.

\section{Detection Model and Dataset}

\subsection{Target Detector}
This study targets a vehicle detection model designed for aerial imagery. The detector is based on YOLOv3~\cite{ref-yolov3}, which is a widely used one-stage object detector known for its efficiency and strong detection performance. YOLOv3 predicts bounding boxes and objectness scores directly from multi-scale feature maps, making it suitable for detecting small objects such as vehicles in aerial images.

YOLOv3 is selected as the target detector for three main reasons. First, its one-stage detection structure provides efficient inference, which is important for aerial monitoring scenarios where a large number of image frames must be processed. Second, YOLOv3 explicitly predicts objectness scores, which directly align with the objective of this study: suppressing the detector's confidence in the presence of vehicles. This makes the relationship between the optimization loss and the AORR/OSR evaluation metrics transparent and interpretable. Third, although YOLOv3 is not the most recent detector, its explicit objectness-based detection formulation makes it particularly suitable for analyzing adversarial patch optimization and digital-to-physical transferability in the considered aerial vehicle detection setting.

To construct the target detector, the model is first pre-trained on the Microsoft COCO dataset~\cite{ref-coco}, which contains diverse object categories and large-scale visual data. This pre-training stage enables the detector to learn general visual features for object recognition. The detector is then fine-tuned for aerial vehicle detection using the Cars Overhead With Context (COWC) dataset~\cite{ref-cowc}. After fine-tuning, the trained detector is used as the target model for evaluating adversarial patch attacks in both the digital and physical domains.

\subsection{Training Dataset}
The target detector is trained using a two-stage dataset configuration consisting of a general-purpose pre-training dataset and an aerial-domain fine-tuning dataset. First, the detector is pre-trained on the Microsoft COCO dataset~\cite{ref-coco}, which is a large-scale benchmark for object detection containing a wide variety of object classes and scene types. This stage provides the detector with general feature representations that are useful for downstream detection tasks.

After pre-training, the detector is fine-tuned on the COWC dataset~\cite{ref-cowc}, which is specifically designed for vehicle detection in overhead imagery. The COWC dataset contains more than 25{,}000 color image patches captured from aerial viewpoints in diverse urban environments. Each image has a spatial resolution of $256 \times 256$ pixels and includes vehicle annotations for detection tasks. Since the dataset focuses on overhead vehicle imagery, it is well suited for adapting the detector to the aerial detection setting considered in this study.

\begin{figure}[!t]
\centering
\subfloat[\centering COWC--Selwyn]{%
    \includegraphics[width=0.28\textwidth]{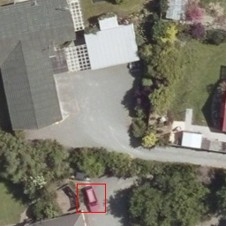}}
\hspace{0.04\textwidth}
\subfloat[\centering COWC--Utah]{%
    \includegraphics[width=0.28\textwidth]{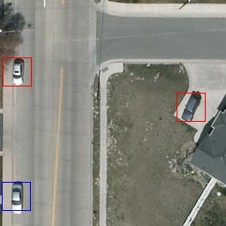}}\\[1.2em]

\subfloat[\centering Car Park]{%
    \includegraphics[width=0.28\textwidth]{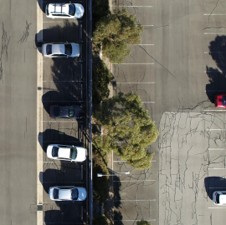}}
\hspace{0.04\textwidth}
\subfloat[\centering Side Street]{%
    \includegraphics[width=0.28\textwidth]{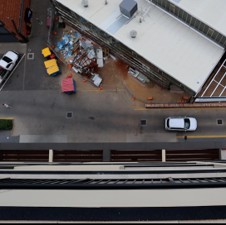}}

\caption{Example images from the datasets used in this study: (\textbf{a}) COWC--Selwyn, (\textbf{b}) COWC--Utah, (\textbf{c}) Car Park, and (\textbf{d}) Side Street.}
\label{fig:dataset_examples}
\end{figure}

Examples from the training data are shown in Figure~\ref{fig:dataset_examples}, including representative scenes from the COWC dataset such as Selwyn and Utah. These images illustrate the overhead viewpoint, small object scale, and scene complexity that characterize aerial vehicle detection.

\subsection{Physical Evaluation Dataset}
To evaluate the real-world effectiveness of adversarial patches, additional aerial images are collected in two physical scenarios: \textit{Side Street} and \textit{Car Park}. These datasets are used exclusively for physical-domain evaluation and are intended to assess whether digitally optimized adversarial patches remain effective after physical deployment.

In the Side Street scenario, aerial images are captured from approximately the height of a ten-story building, corresponding to an observation height of about 40~m. In the Car Park scenario, aerial images are collected using an unmanned aerial vehicle (UAV) flying at an altitude of approximately 60~m. These two settings provide different viewpoints and scene structures, allowing the robustness of the adversarial patches to be evaluated under varying environmental conditions.

During physical data collection, three vehicles under our control are used, namely gray, white, and blue vehicles, depending on the scenario. The images and videos are recorded under multiple environmental conditions, including direct sunlight and shade, as well as static and moving vehicle states. The resulting physical evaluation dataset consists of 5{,}225 frames for the Side Street scenario and 2{,}126 frames for the Car Park scenario. Example images from these physical evaluation environments are also shown in Figure~\ref{fig:dataset_examples}.

These datasets enable a comprehensive evaluation of adversarial patch transferability from digital optimization to physical deployment under realistic aerial capture conditions.

\section{Experimental Setup}

All experiments in this study were conducted on a workstation equipped with an NVIDIA GeForce RTX 4070 GPU. The models were implemented and trained using PyTorch 2.5.1 with CUDA 12.1.

\subsection{Attack Settings}
To evaluate the effectiveness of adversarial patches against aerial vehicle detectors, the patches are first optimized in the digital domain and then deployed in the physical domain. During optimization, each patch is iteratively updated to minimize the objectness score predicted by the target detector while satisfying the printability and smoothness constraints described in Equation~(\ref{eq:patch_loss}). The optimization objective combines the maximum objectness term with the Non-Printability Score (NPS) and Total Variation (TV) regularization in order to generate adversarial patterns that are both physically realizable and robust under real-world capture conditions.

During patch training, several transformations are applied to improve robustness against environmental variations. Specifically, geometric transformations such as random scaling and rotation are used to simulate viewpoint changes in aerial imagery. In addition, color-space transformations including brightness and contrast adjustments are introduced to account for illumination variations. Weather-based augmentation is also incorporated to approximate environmental disturbances such as atmospheric effects and lighting variation.

To analyze the effect of augmentation strategies during patch optimization, three training pipelines are considered in this study: \textit{G/C+W}, \textit{G/C}, and \textit{Control}. The \textit{G/C+W} setting includes geometric transformations, color-space transformations, and weather augmentation. The \textit{G/C} setting includes only geometric and color-space transformations. The \textit{Control} setting uses a random patch without adversarial optimization. In addition, three patch placement strategies are evaluated, namely the ON, OFF, and OFF-Side configurations described in the previous section. These settings allow us to compare how augmentation strategies and patch placement affect attack performance in both digital and physical environments.

\subsection{Weather Augmentation}
To evaluate the robustness of adversarial patches under environmental appearance changes, image-level weather augmentation is applied during both patch training and digital testing when weather augmentation is enabled. In this process, the adversarial patch is first geometrically transformed according to the target object location and pasted onto the clean image. Then, the weather transformation is applied to the entire patched image before resizing it to the YOLOv3 input resolution. Therefore, the weather effect influences both the background scene and the adversarial patch, which better approximates physical deployment conditions where illumination, rain, fog, and color shifts affect the whole captured image.

\begin{figure}[!t]
\centering
\subfloat[\centering Brighten]{%
    \includegraphics[width=0.22\textwidth]{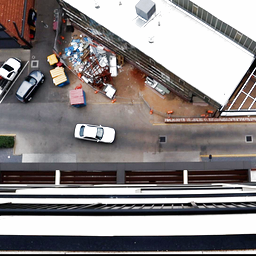}}
\hspace{0.02\textwidth}
\subfloat[\centering Darken]{%
    \includegraphics[width=0.22\textwidth]{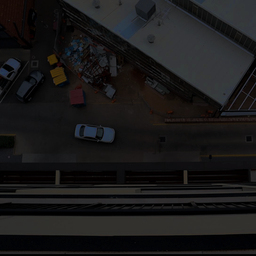}}
\hspace{0.02\textwidth}
\subfloat[\centering Snow]{%
    \includegraphics[width=0.22\textwidth]{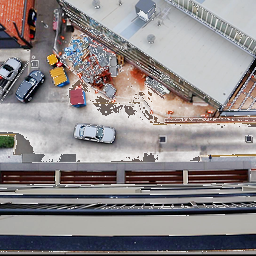}}
\hspace{0.02\textwidth}
\subfloat[\centering Rain]{%
    \includegraphics[width=0.22\textwidth]{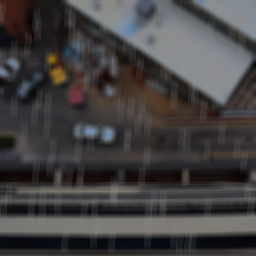}}\\[1.0em]

\subfloat[\centering Fog]{%
    \includegraphics[width=0.22\textwidth]{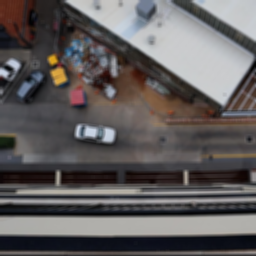}}
\hspace{0.02\textwidth}
\subfloat[\centering Autumn]{%
    \includegraphics[width=0.22\textwidth]{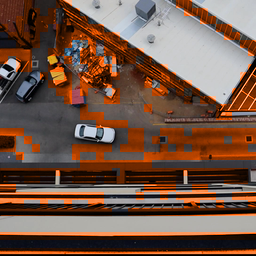}}
\hspace{0.02\textwidth}
\subfloat[\centering Identity]{%
    \includegraphics[width=0.22\textwidth]{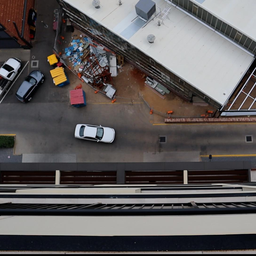}}
\caption{Examples of the seven image-level weather augmentation cases used in patch training and digital testing: (\textbf{a}) brighten, (\textbf{b}) darken, (\textbf{c}) snow, (\textbf{d}) rain, (\textbf{e}) fog, (\textbf{f}) autumn color shift, and (\textbf{g}) identity transformation.}
\label{fig:weather_examples}
\end{figure}

For each image, one weather condition is randomly selected with uniform probability from seven candidates: brightening, darkening, snow, rain, fog, autumn color shift, and identity transformation. This selection corresponds to sampling an integer value from 0 to 6, where the seven cases denote brightening, darkening, snow, rain, fog, autumn color shift, and no transformation, respectively. The same weather augmentation policy is used during digital testing under the \textit{STD+W} setting. Representative examples of the seven weather augmentation cases are shown in Figure~\ref{fig:weather_examples}.

The brightening and darkening transformations are performed in the HLS color space by modifying the lightness channel. For brightening, the lightness coefficient is sampled as $1+\mathcal{U}(0,1)$, resulting in a scaling factor in the range $[1.0,2.0]$. For darkening, the coefficient is sampled as $1-\mathcal{U}(0,1)$, producing a scaling factor in the range $[0.0,1.0]$. The snow transformation increases the lightness of relatively dark pixels. Its threshold is sampled as $\mathcal{U}(0,1)\times 255/2 + 255/3$, corresponding approximately to the range $[85,212]$, and pixels below this threshold are brightened using a fixed coefficient of 2.5.

The rain transformation generates random rain streaks over the image. The slant of the streaks is selected from the integer range $[-10,10]$, while the default drop length, drop width, and drop color are set to 20, 1, and $(200,200,200)$, respectively. The number of rain drops is proportional to the image area and is computed as the image area divided by 600. After drawing the rain streaks, a blur operation with a $7 \times 7$ kernel is applied, and the lightness channel is multiplied by 0.7 to simulate darker rainy conditions. The fog transformation overlays semi-transparent white circular haze patterns, with the fog coefficient sampled from $\mathcal{U}(0.3,1.0)$. The haze size is determined as one third of the image width multiplied by the fog coefficient, and the final image is blurred using a kernel size proportional to the haze size. Finally, the autumn transformation divides the image into local $8 \times 8$ regions and changes the hue of selected low-green regions to one of four autumn-like hue values, $\{1,5,9,11\}$, thereby simulating seasonal background color shifts.

Overall, the processing order for weather-based experiments is as follows: the clean image is padded to a square shape, the transformed adversarial patch is pasted, weather augmentation is applied to the patched image, the image is resized to $256 \times 256$, and YOLOv3 inference is then performed. For non-weather experiments, the same procedure is used except that the weather augmentation step is skipped.

\subsection{Evaluation Metrics}
To quantitatively evaluate the effectiveness of adversarial patch attacks, this study adopts objectness-based metrics for both digital-domain and physical-domain experiments.
In the digital domain, the attack performance is measured using the Average Objectness Reduction Rate (AORR), which quantifies the relative decrease in objectness score after applying the adversarial patch. AORR is defined as

\begin{equation}
\mathrm{AORR}(T^{v}, \hat{T}^{v}) \triangleq \frac{1}{N E_j} \sum_{j=1}^{N} \sum_{l=1}^{E_j} \frac{s_{j,l}^{v} - \hat{s}_{j,l}^{v}}{s_{j,l}^{v}},
\label{eq:aorr}
\end{equation}

\noindent where $T^{v}$ denotes the set of detection results for all images in the test set $v$, $N$ is the number of test images, and $E_j$ represents the number of detected objects in the $j$-th image. The term $s_{j,l}^{v}$ denotes the original objectness score of the $l$-th detection in the $j$-th image, whereas $\hat{s}_{j,l}^{v}$ denotes the corresponding objectness score after applying the adversarial patch. A higher AORR value indicates a stronger attack effect because it reflects a larger reduction in the detector's confidence.

We use AORR because the adversarial patch is optimized to directly suppress the objectness score of the target detector. Therefore, AORR measures the relative change in the same confidence signal that is minimized during patch optimization and provides a direct view of whether the learned patch successfully weakens the detector's vehicle-related response. Standard detection metrics such as mAP remain important for assessing final detection performance, but they can be affected by thresholding, non-maximum suppression, and localization changes. In contrast, AORR captures confidence degradation before and after patch application at the objectness score level. For this reason, we report AORR as the primary attack-oriented metric and additionally report mAP in the digital-domain evaluation to verify whether objectness suppression also leads to actual detection-performance degradation.

For physical-domain evaluation, we use the Objectness Score Ratio (OSR), which measures the relative reduction in detection confidence after deploying the printed adversarial patch. OSR is defined as

\begin{equation}
  \mathrm{OSR}(\hat{S}, S) \triangleq
  \frac{\frac{1}{\alpha}\sum_{f=1}^{\alpha} \hat{S}_f}
  {\frac{1}{\beta}\sum_{g=1}^{\beta} S_g},
  \label{eq:osr}
\end{equation}

\noindent where $S_g$ denotes the original objectness score of the target object, and $\hat{S}_f$ represents the objectness score after the adversarial patch is applied. The parameter $\alpha$ denotes the total number of frames captured when the adversarial patch is physically deployed, whereas $\beta$ denotes the total number of frames captured without the adversarial patch. A lower OSR value indicates a more effective physical attack.

OSR is used in the physical-domain evaluation because perfectly paired clean and patched images are difficult to obtain in real aerial capture. Unlike the digital setting, where the same image can be evaluated before and after inserting a patch, physical experiments are affected by small changes in viewpoint, illumination, vehicle position, camera motion, and environmental conditions. Therefore, directly computing an image-wise objectness reduction rate analogous to AORR is not reliable in the physical domain. Instead, OSR compares the average objectness score of frames captured with the printed patch to that of frames captured without the patch, providing a practical measure of confidence reduction under real deployment conditions.

\subsection{Digital-Domain Evaluation}
In the digital-domain experiments, adversarial patches are directly inserted into test images to evaluate their effectiveness without physical deployment. The experiments are conducted using the aerial vehicle detector described in the previous section, which is based on the YOLOv3 architecture.

To evaluate the robustness of the optimized patches under different testing conditions, two testing regimes are considered. The first regime, denoted as \textit{STD}, measures AORR on the original test dataset without additional perturbation. The second regime, denoted as \textit{STD+W}, applies weather effects to the test images before evaluation. These two settings allow us to analyze whether the optimized patches remain effective when environmental variations are introduced during testing.

The digital-domain evaluation is performed on the Side Street dataset. We do not conduct the same digital analysis on the Car Park dataset because vehicles in that environment are parked very close to one another, causing visual interference among multiple patches and making it difficult to isolate the effect of each patch configuration. For each training pipeline and patch configuration, the attack performance is measured using AORR under both the \textit{STD} and \textit{STD+W} settings.

\subsection{Physical-Domain Evaluation}
To evaluate real-world attack effectiveness, the optimized adversarial patches are printed and deployed in physical environments. The physical experiments are conducted in two scenarios: \textit{Side Street} and \textit{Car Park}. In the Side Street scenario, images are captured from approximately 40~m above the ground, corresponding to the height of a ten-story building. In the Car Park scenario, aerial images are collected using a UAV flying at an altitude of approximately 60~m.

For physical deployment, the adversarial patches are printed on 160~gsm coated paper and placed either on the roof of the vehicle (ON patch) or around the vehicle (OFF patch). The OFF-Side configuration is considered in digital-domain analysis, whereas the physical-domain experiments mainly focus on the printed ON and OFF configurations. During data collection, the vehicles are recorded under varying environmental conditions, including direct sunlight and shade, as well as static and moving states.

The captured aerial images and video frames are processed by the target detector to obtain objectness scores for the target vehicle. The effectiveness of the adversarial attack is then measured using the OSR metric defined in Equation~(\ref{eq:osr}). Through this evaluation protocol, we assess whether digitally optimized adversarial patches can maintain their effectiveness after physical printing, placement, and aerial image capture.

\section{Results and Analysis}

\subsection{Digital-Domain Results}
Digital-domain evaluation reports results only on the Side Street dataset. We omit analyses on the Car Park dataset because vehicles in that environment are parked very close to one another, causing multiple patches to visually interfere with each other. This inter-patch interference makes it difficult to isolate the effect of each patch configuration and may distort the evaluation of attack effectiveness.

\begin{table*}[htbp]
\centering
\caption{Digital-domain evaluation results of adversarial patches on the Side Street dataset under both STD and STD+W evaluation settings. For patched cases, bold and underline indicate the best and second-best attack results, respectively; lower mAP and higher AORR indicate stronger attacks.}
\label{tab:digital_results}
\small
\begin{tabular}{c l c c c c}
\toprule
\makecell{\textbf{Training}\\\textbf{pipeline}} &
\makecell{\textbf{Patch}\\\textbf{type}} &
\makecell{\textbf{mAP}\\\textbf{(STD) [\%]}} &
\makecell{\textbf{AORR}\\\textbf{(STD) [\%]}} &
\makecell{\textbf{mAP}\\\textbf{(STD+W) [\%]}} &
\makecell{\textbf{AORR}\\\textbf{(STD+W) [\%]}} \\
\midrule
-- & Clean &  56.22 & -- & 34.15 & -- \\
\midrule
G/C & ON &  5.98 & 72.59 & 7.07 & 64.09 \\
G/C+W & ON  & \textbf{4.10} & 53.76 & \textbf{6.38} & 54.44 \\
Control & ON &  20.01 & 29.06 & 16.48 & 34.75 \\
\midrule
G/C & OFF & 7.46 & \textbf{85.51} & 9.52 & \textbf{78.02} \\
G/C+W & OFF  & \underline{4.93} & \underline{79.37} & \underline{6.57} & \underline{71.80} \\
Control & OFF & 27.28 & 20.65 & 18.01 & 10.73 \\
\midrule
G/C & OFF-Side  & 10.85 & 67.43 & 13.50 & 62.43 \\
G/C+W & OFF-Side  & 5.54 & 66.46 & 12.28 & 59.36 \\
Control & OFF-Side & 39.65 & 15.66 & 28.86 & 28.49 \\
\bottomrule
\end{tabular}
\end{table*}

Table~\ref{tab:digital_results} summarizes the AORR results for the three patch configurations under different training pipelines. Overall, the \textit{G/C} pipeline consistently outperforms the \textit{G/C+W} pipeline for all patch types under both the \textit{STD} and \textit{STD+W} evaluation settings. This result suggests that incorporating weather-based augmentation does not improve adversarial patch optimization in this setting and may even degrade attack effectiveness. In other words, weather augmentation appears to be unnecessary for generating effective adversarial patches for aerial vehicle detection.

Out of the three patch configurations, the OFF patch achieves the strongest overall attack performance in the digital domain. In particular, the OFF patch trained with the \textit{G/C} pipeline attains the highest AORR values, reaching 85.51\% under \textit{STD} and 78.02\% under \textit{STD+W}. These results indicate that perturbing the contextual region around the vehicle is more effective in the digital domain than directly modifying the appearance of the vehicle itself.

The OFF-Side patch also exhibits meaningful attack performance, but it does not surpass the OFF patch. Specifically, the OFF-Side patch achieves 67.43\% under \textit{STD} and 62.43\% under \textit{STD+W} with the \textit{G/C} pipeline, which is higher than the ON patch in some cases but still lower than the OFF patch overall. These results suggest that attack effectiveness varies depending on the number and spatial arrangement of the deployed patches.

Table~\ref{tab:digital_results} also shows that applying adversarial patches substantially decreases the detection accuracy compared with the clean baseline. Under the \textit{STD} setting, the clean image achieves 56.22\% mAP, whereas the adversarially optimized patches reduce the mAP to 5.98\%, 7.46\%, and 10.85\% for the ON, OFF, and OFF-Side configurations with the \textit{G/C} pipeline, respectively. A similar degradation is observed under the \textit{STD+W} setting, where the clean mAP is 34.15\%, but the adversarial patch cases remain much lower, ranging from 6.38\% to 13.50\% depending on the patch configuration and training pipeline. These mAP reductions are consistent with the AORR results: patch configurations with large objectness reduction generally produce severe detection-performance degradation. For example, the OFF patch with the \textit{G/C} pipeline achieves the highest AORR under both \textit{STD} and \textit{STD+W}, while also reducing mAP far below the clean baseline. This consistency indicates that the AORR metric reflects not only confidence suppression but also practical degradation in detection performance. In contrast, the control patches generally retain higher mAP values or exhibit lower AORR values than the adversarially optimized patches, indicating that the performance drop is mainly caused by the optimized adversarial patterns rather than by simple occlusion or patch placement alone.

Overall, the digital-domain results reveal three key findings: (1) weather-based augmentation is not beneficial for patch optimization in this setting, (2) the OFF patch is the most effective configuration in the digital domain, and (3) attack performance changes according to the number and spatial extent of the deployed patches.

\subsection{Physical-Domain Results}
In the physical-domain evaluation, the optimized adversarial patches are printed on 160~gsm coated paper, and aerial videos are recorded at 25~FPS to construct the test dataset for both the Car Park and Side Street scenarios. During data collection, three vehicles under our control, namely gray, white, and blue vehicles, are equipped with the printed adversarial patches. The resulting dataset consists of 2{,}126 frames for the Car Park scenario and 5{,}225 frames for the Side Street scenario.

\begin{table}[htbp]
\centering
\caption{Results of adversarial patches in the physical domain~\cite{ref-du}.}
\label{tab:physical_results}
\begin{tabular}{l l l l c}
\toprule
\multicolumn{5}{c}{\textbf{(a) Side Street}} \\
\midrule
\textbf{Car} & \textbf{Patch type} & \textbf{Lighting} & \textbf{Motion} & \textbf{Mean OSR} \\
\midrule
Gray  & ON  & Both  & Moving & 0.343 \\
Gray  & ON  & Sun   & Static & 0.251 \\
Gray  & ON  & Shade & Static & 0.255 \\
Gray  & OFF & Sun   & Static & 0.429 \\
\midrule
White & ON  & Both  & Moving & 0.286 \\
White & ON  & Sun   & Static & 0.285 \\
White & ON  & Shade & Static & 0.197 \\
White & OFF & Sun   & Static & 0.748 \\
\midrule
\multicolumn{5}{c}{\textbf{(b) Car Park}} \\
\midrule
\textbf{Car} & \textbf{Patch type} & \textbf{Lighting} & \textbf{Motion} & \textbf{Mean OSR} \\
\midrule
Gray  & ON & Sun   & Static & 0.509 \\
Blue  & ON & Shade & Static & 0.208 \\
White & ON & Sun   & Static & 0.746 \\
\bottomrule
\end{tabular}
\end{table}

Table~\ref{tab:physical_results} summarizes the physical-domain attack results. Overall, the deployed patches significantly reduce the detector's objectness scores across different vehicles and environmental conditions. The measured OSR values range from 0.197 to 0.746, indicating substantial degradation in detection confidence after physical patch deployment.

Unlike the digital-domain results, where the OFF patch achieves the best performance, the ON patch generally shows stronger effectiveness in the physical domain. In the Side Street scenario, the ON patch reduces the mean OSR to values between 0.197 and 0.343 depending on the vehicle, lighting condition, and motion state. In contrast, the OFF patch shows weaker performance in the same scenario, with OSR values of 0.429 and 0.748 in the evaluated settings. In the Car Park scenario, the ON patch also reduces the detector confidence, achieving OSR values of 0.208, 0.509, and 0.746 for different vehicles and lighting conditions.

These results show that adversarial patches optimized in the digital domain can be transferred to real-world aerial environments and remain effective after physical deployment. At the same time, the results indicate that the most effective patch configuration differs between the digital and physical domains.

\subsection{Effect of TV Regularization}
\begin{figure}[htbp]
\centering
\subfloat[\centering $\gamma = 1.0$]{%
    \includegraphics[width=0.30\textwidth]{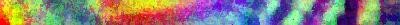}}
\hspace{0.03\textwidth}
\subfloat[\centering $\gamma = 5.5$]{%
    \includegraphics[width=0.30\textwidth]{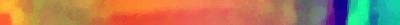}}
\hspace{0.03\textwidth}
\subfloat[\centering $\gamma = 40.0$]{%
    \includegraphics[width=0.30\textwidth]{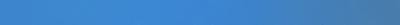}}

\caption{Visual examples of optimized patches under different TV loss coefficients: (\textbf{a}) $\gamma = 1.0$, (\textbf{b}) $\gamma = 5.5$, and (\textbf{c}) $\gamma = 40.0$.}
\label{fig:tv_patches}
\end{figure}

To analyze the effect of the TV regularization term, we vary the coefficient $\gamma$ in the patch optimization loss and evaluate the OFF patch under the \textit{G/C+W} training pipeline. The results are summarized in Table~\ref{tab:tv_results}, and representative visual examples of the optimized patches are shown in Figure~\ref{fig:tv_patches}.

Overall, the attack performance strongly depends on the value of the TV loss coefficient. When $\gamma = 0.0$, the AORR is relatively low, reaching 33.18\% under \textit{STD} and 36.76\% under \textit{STD+W}, indicating that patches without smoothness regularization are less effective. Introducing TV regularization significantly improves the attack performance, with the best results observed around $\gamma = 1.0$ and $\gamma = 2.5$. Specifically, the highest AORR values are obtained at $\gamma = 1.0$ under \textit{STD} and at $\gamma = 2.5$ under \textit{STD+W}.

\begin{table}[htbp]
  \centering
\caption{Results of adversarial patches under the G/C+W training pipeline with different TV loss coefficients. Bold and underline indicate the best and second-best attack results, respectively.}
  \label{tab:tv_results}
  \begin{tabular}{c c c}
	  \toprule
	  \textbf{TV loss coefficient ($\gamma$)} & \textbf{AORR (STD) [\%]} & \textbf{AORR (STD+W) [\%]} \\
	  \midrule
	  0.0  & 33.18 & 36.76 \\
	  1.0  & \textbf{80.79} & \underline{71.32} \\
	  2.5  & \underline{79.37} & \textbf{71.80} \\
	  4.0  & 76.23 & 69.17 \\
	  5.5  & 72.54 & 66.72 \\
	  20.0 & 53.52 & 56.26 \\
	  40.0 & 47.19 & 53.23 \\
	  \bottomrule
  \end{tabular}
\end{table}

However, as the value of $\gamma$ increases further, the attack performance gradually decreases. For example, when $\gamma = 20.0$ or $\gamma = 40.0$, the AORR drops substantially in both evaluation settings. This trend indicates that excessive TV regularization overly smooths the patch, thereby weakening the adversarial patterns that are necessary for suppressing the detector's objectness score.

The visual examples in Figure~\ref{fig:tv_patches} support this observation. When $\gamma$ is small, the optimized patch contains diverse and high-frequency color patterns that effectively disrupt the detector. As $\gamma$ increases, the patch becomes progressively smoother, and at very large values, such as $\gamma = 40.0$, the patch converges to a nearly uniform color pattern. These results indicate that moderate TV regularization is necessary to balance printability and attack effectiveness, whereas excessive smoothing degrades adversarial performance.

\section{Discussion}
This study focuses on evaluating the digital-to-physical transfer of adversarial patches in aerial vehicle detection, where attacks are affected by overhead viewpoints, small object scale, long-distance capture, physical visibility, and context dependence. These factors distinguish the aerial setting from classic ground-based physical patch scenarios.

The experimental results provide several important observations regarding the transferability of adversarial patches from digital optimization to physical deployment in aerial vehicle detection.  First, the digital-domain results show that the OFF patch achieves the strongest attack performance, whereas the physical-domain results indicate that the ON patch is generally more effective after real-world deployment. This discrepancy suggests that the attack effectiveness in the digital domain does not necessarily translate directly to the physical domain. One possible explanation for this difference is the distinct role of contextual perturbations and object-level perturbations under real-world imaging conditions. In the digital domain, the OFF patch is highly effective because it perturbs contextual cues around the vehicle, such as road texture, parking-lot structure, shadows, and local background contrast, while its position, scale, and visibility are precisely controlled. However, this context-dependent attack becomes less stable after physical deployment, where viewpoint changes, partial occlusion, scale variation, boundary misalignment, environmental noise, and small placement errors can alter the spatial relationship between the vehicle and the surrounding patch. In contrast, the ON patch is attached directly to the vehicle body and moves together with the target object, maintaining more consistent visibility and object-level alignment across aerial frames. This explains why the ON patch is more robust in the physical domain, even though the OFF patch is stronger in digital evaluation. The OFF-Side configuration further shows the role of spatial coverage, as perturbing only the two side regions provides weaker contextual manipulation than the full OFF configuration. Since OFF-Side is mainly evaluated digitally in this study, a complete physical comparison among ON, OFF, and OFF-Side remains future work.

Second, the results show that weather-based augmentation during optimization does not improve attack performance and may even reduce it. In the digital-domain experiments, the \textit{G/C} pipeline consistently outperforms the \textit{G/C+W} pipeline for all patch configurations. This finding suggests that the additional variability introduced by weather augmentation might hinder the optimization of discriminative adversarial patterns in the considered setting. Rather than improving robustness, excessive augmentation may weaken the attack objective by forcing the patch to adapt to overly broad or unrealistic perturbation conditions. A possible mechanism behind this result is that the weather transformation perturbs both the adversarial patch and the surrounding scene after patch placement. Because the attack objective is based on suppressing objectness scores, the optimizer must learn patch patterns that consistently disturb vehicle-related features across many transformed appearances. Strong image-level weather effects such as fog, rain, snow, and large brightness changes can alter edge contrast, local texture statistics, and color distributions of the entire image. As a result, gradients obtained from different weather conditions may become less consistent across iterations, making the optimization less stable and reducing the emergence of sharp, discriminative adversarial patterns. In addition, weather augmentation may partially mask or smooth the patch itself, especially when haze, rain blur, or strong illumination changes are applied after patch insertion. This can weaken the feature perturbation produced by the patch and encourage the optimization to converge to patterns that are more generic but less effective against the detector's stable vehicle features. Therefore, in this setting, weather augmentation appears to increase appearance diversity but does not necessarily improve adversarial generalization. Instead, it can dilute the attack signal and reduce optimization efficiency.

Third, the ablation study on TV regularization demonstrates that moderate smoothness constraints are essential for effective physical adversarial patch generation. When the TV coefficient is too small, the optimized patch lacks sufficient structural regularity for stable physical deployment. On the other hand, when the coefficient becomes too large, the patch becomes overly smooth and loses the high-frequency patterns necessary for suppressing detector confidence. This result indicates that there is a trade-off between printability and adversarial effectiveness, and that an intermediate level of regularization provides the best balance.

Overall, these findings highlight several practical implications for aerial object detection security. They show that aerial detectors are vulnerable not only to digitally optimized perturbations but also to physically deployed patches that remain effective under real-world capture conditions. At the same time, the results reveal that the most effective attack strategy depends strongly on whether the evaluation is conducted in the digital or physical domain. Therefore, robustness assessment for aerial object detectors should not rely solely on digital experiments, but should also consider physical deployment scenarios and environmental variability.

These observations also suggest several defense implications. First, robustness evaluation and adversarial training for aerial detectors should include physically meaningful patch placements rather than relying only on pixel-level perturbations, because ON and OFF patches affect object-level and contextual features in different ways. Second, since the OFF patch is effective in the digital domain by perturbing contextual cues, detectors should be encouraged to rely less on fragile background context and more on stable vehicle shape and structural features. This may be supported by context-randomized training, patch-aware augmentation, or feature-level regularization. Third, because the ON patch remains effective in physical environments due to consistent visibility, defense methods should consider localized abnormal-pattern detection on vehicle surfaces and temporal consistency checks in aerial video. A patch that suppresses objectness in individual frames may still produce spatially or temporally unusual responses across consecutive frames. These directions are consistent with recent adversarial patch defense studies in general object detection, which emphasize both patch localization/removal and rigorous defense evaluation. Jing et al. \cite{ref-pad} proposed a patch-agnostic defense that localizes and removes adversarial patches without prior attack knowledge, while Zheng et al. \cite{ref-apde} further showed, through a unified benchmark, that patch localization accuracy alone may not reliably reflect defense effectiveness and that object-level detection metrics such as AP and ASR should also be considered. Finally, confidence calibration and ensemble-based detection can be used to reduce overreliance on a single objectness signal. Although these defenses are beyond the scope of the present attack-focused study, they provide practical directions for improving aerial detector robustness against physical adversarial perturbations.

Despite these findings, this study has limitations. First, the experiments are conducted using a single detector architecture, YOLOv3, and the generality of the results to other aerial object detectors remains to be verified. Second, the physical evaluation is limited to a small number of vehicles and capture environments. Third, the OFF-Side configuration is mainly evaluated in the digital domain, and further physical experiments would provide a more complete comparison among all patch types. In addition, the current weather analysis does not explicitly consider dynamic shadows or partial patch occlusion caused by buildings, trees, or changes in sun position, although such effects are important in real aerial photography. Future work should extend the analysis to additional detector architectures, more diverse aerial scenarios, dynamic shadow and partial-overlap conditions, and defense strategies for improving robustness against physical adversarial perturbations.

\section{Conclusion}
This paper investigated the transferability of adversarial patches from digital optimization to physical deployment for aerial vehicle detection. Adversarial patches were optimized in the digital domain using an objectness-based loss with Non-Printability Score and Total Variation regularization, and were then evaluated under three placement strategies: ON, OFF, and OFF-Side.

The experimental results showed that the OFF patch achieved the strongest attack effectiveness in the digital domain, indicating that contextual perturbations around the vehicle can strongly suppress detector confidence in simulation-based evaluation. In contrast, the ON patch was more effective in the physical domain, suggesting that object-level perturbations are more robust after printing and real-world deployment. In addition, weather-based augmentation did not improve patch optimization performance in our setting, and the TV loss coefficient was found to have a substantial influence on attack effectiveness.

These results demonstrate that digitally optimized adversarial patches can transfer to real-world aerial environments and significantly degrade the performance of aerial vehicle detectors. More importantly, the study reveals that the most effective patch configuration differs between digital and physical evaluations. This finding highlights the importance of considering physical deployment conditions when assessing the robustness of aerial object detection systems.

Overall, this work provides empirical evidence of the practical vulnerability of aerial object detectors to physical adversarial attacks and offers useful insights for future research on robust aerial perception and adversarial defense.


\end{document}